\documentclass[10pt,twocolumn,letterpaper]{article}
\usepackage[pagenumbers]{cvpr} 

\usepackage{graphicx}
\usepackage{array}
\usepackage{rotating}

\usepackage[dvipsnames]{xcolor}

\definecolor{cvprblue}{rgb}{0.21,0.49,0.74}
\usepackage[pagebackref,breaklinks,colorlinks,citecolor=cvprblue]{hyperref}

\def\paperID{EqVis4} 
\def\confName{CVPR}
\def\confYear{2024}

\title{SERNet-Former: Semantic Segmentation by Efficient Residual Network with Attention-Boosting Gates and Attention-Fusion Networks}

\author{Serdar Erişen\\
Hacettepe University\\
{\tt\small serdarerisen@hacettepe.edu.tr}
}

\begin{document}
\maketitle
\begin{abstract}
Improving the efficiency of state-of-the-art methods in semantic segmentation requires overcoming the increasing computational cost as well as issues such as fusing semantic information from global and local contexts. Based on the recent success and problems that convolutional neural networks (CNNs) encounter in semantic segmentation, this research proposes an encoder-decoder architecture with a unique efficient residual network, Efficient-ResNet. Attention-boosting gates (AbGs) and attention-boosting modules (AbMs) are deployed by aiming to fuse the equivariant and feature-based semantic information with the equivalent sizes of the output of global context of the efficient residual network in the encoder. Respectively, the decoder network is developed with the additional attention-fusion networks (AfNs) inspired by AbM. AfNs are designed to improve the efficiency in the one-to-one conversion of the semantic information by deploying additional convolution layers in the decoder part. Our network is tested on the challenging CamVid and Cityscapes datasets, and the proposed methods reveal significant improvements on the residual networks. To the best of our knowledge, the developed network, SERNet-Former, achieves state-of-the-art results (84.62 \% mean IoU) on CamVid dataset and challenging results (87.35 \% mean IoU) on Cityscapes validation dataset. Please also visit the project page: \url{https://github.com/serdarch/SERNet-Former}
\end{abstract}    
\section{Introduction}
\label{sec:intro}

Semantic segmentation is a fundamental computational
task that is widely applied in 2D scene understanding in the research field of computer vision. In semantic
segmentation, each pixel is mapped through the labeled
semantic classes, introduced via the ground truth of an
image input, that is tried to be predicted by state-of-the-art
networks and methods. Among the number of benefits of
semantic segmentation, autonomous driving, and robotics
in recognizing indoor and outdoor scenes, medical
imaging, virtual reality, augmented reality, real-time
surveillance, scene understanding, photography, creating,
and editing images \cite{r1, r2, r3} can be counted as the widely used
applications and emerging research fields. Many different
types of deep neural networks (DNNs), including Fully
Connected Network (FCN) and Convolutional Neural
Networks (CNNs), applying encoder-decoder architectures
and attention-based models, have shown remarkable
progress \cite{r2, r3, r4}. Recently, Vision Transformers (ViT) with
Swin Transformers \cite{r5, r6, r7, r8} and CNN architectures achieved
state-of-the-art performances in semantic segmentation \cite{r2, r4}.

\begin{figure}
    \centering    
    \includegraphics[width=1\linewidth]{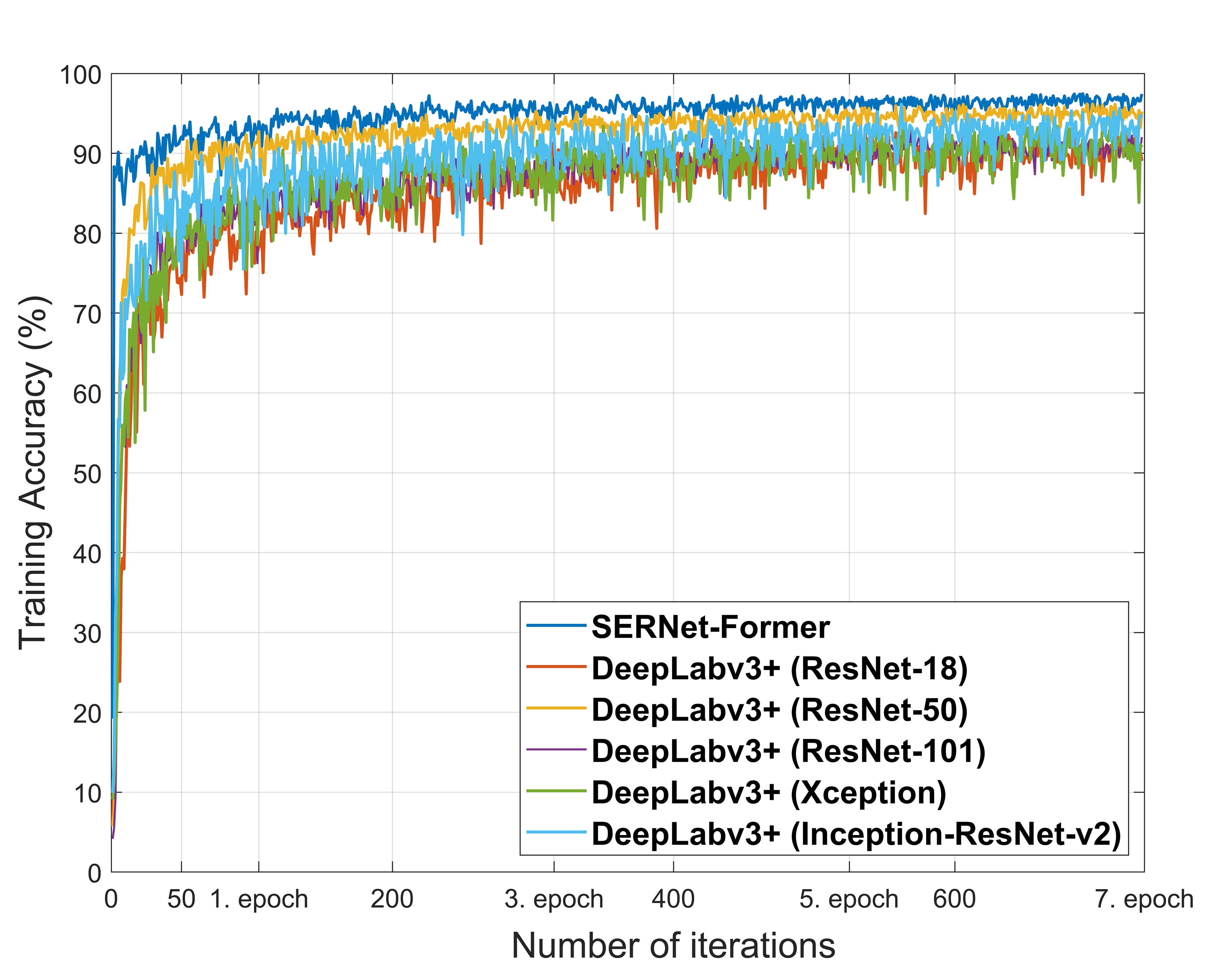}
    \caption{Training progress of SERNet-Former and the selected
baselines of DeepLabv3+ on CamVid dataset.}
\label{fig:1}
\end{figure}

Fusing the multi-scale semantic information is one of the
compelling issues in segmentation tasks besides the
concerns of computational cost and efficiency of the
developed networks and methods. The challenge of
recognizing objects in semantic segmentation in 2D scenes
is two-fold: The labeled object may either lose its spatial
information or feature-rich properties during the
processing of the image throughout the network. Most
recent state-of-the-art networks \cite{r2, r4, r9} also try to
overcome the inconsistency between the semantic
information from the global and local contexts of a given
image input. Thus, the spatial information and feature maps
in CNNs are aimed to be discovered deeply for efficient
and precise semantic segmentation in this work. 

It is another intriguing fact that increasing the number of
convolution layers does not always return efficient results
when compared to its computational cost \cite{r10} (Fig. \ref{fig:1}). In
that regard, rather than using the largest CNNs, the
research dives into the deep potential of the improved
performance of CNNs with additional methods. This work
re-evaluates the efficiency of the CNN-based encoder-decoder architectures with attention-based fusion networks
and modules. The study discovers the potential of fused
attention gates, not only used in the encoder of the
developed network architectures but also fused within the
decoder.

Therefore, it is first explored to find the most efficient
baseline architecture that can be trained fast and should be
lightweight (Fig. \ref{fig:1}). For instance, residual networks \cite{r11}
are analyzed as the baselines for further improvement on
the encoder network. Different types of attention gates are
developed together with convolution layers in the encoder
and decoder parts. The role of spatial information, fused
with channel-based semantic information, is also
discovered by carefully placing the skip connections at the
right steps in the encoder and decoder networks.

Thus, the encoder of our network is improved with the
attention-boosting gates (AbGs), resulting in an efficient
residual network, Efficient-ResNet, with increased training
performance and prediction efficiency. Notwithstanding
this, larger networks can perform far better than the
shallower networks in longer training periods. In that
regard, attention-fusion networks (AfNs) are deployed as
another method, adapted to the decoder part of our
network, to improve its capacity by storing, fusing, and
processing rich semantic information from the encoder
during up-sampling. Thus, attention-fusion networks are
designed to superpose the global spatial context with
feature-based rich semantic information to improve the
performance of CNNs with fewer layers. Our methods
show significant improvements in the trained baseline
architectures in the initial epochs on CamVid dataset \cite{r12, r13}. 
Fig. \ref{fig:1} illustrates the comparison of the initial training
performance of some networks using CNNs as the baseline
architectures. Respectively, SERNet-Former learns much
faster than the quickest learning baseline that is applied in
DeepLabv3+ architecture \cite{r9, r14} during training sessions on CamVid dataset (Fig. \ref{fig:1}).

In brief, our network is developed from the residual
CNNs by adding attention-boosting gates and attention-fusion networks. Our additional methods are connected to
the encoder and decoder parts via skip connections to fuse
the rich information from different contexts, concatenated
by superposing the useful semantic information for the
highest efficiency. To the best of our knowledge, an
‘efficient residual network’ is developed that is unique in
the literature used as the encoder of our network. The
network is improved by attention-boosting gates and
attention-fusion networks with increasing efficiency and
precision in predicting and recognizing smaller objects and
their features. As being highly inspired by RGB-D
networks \cite{r15, r16}, the methods of attention-based fusion modules and networks are also deployed with an aim to
adapt our model for different classification tasks and
feature maps from RGB-D networks and 3D point clouds
for future work. Our contributions can be briefly
highlighted as follows:

\begin{itemize}
        \item A unique ‘efficient residual network’ is developed as an encoder with the additional excitation by attention-boosting gates (AbGs) with a search for the optimal training performance and computational cost of CNNs
        \item The capacity of the decoder part of our network is
        improved via attention-fusion networks (AfNs),
        increasing the efficiency of acquiring and processing
        the feature-rich semantic information
        \item Skip connections, turning the decoder part into a
        superposition network, are designed to fuse and
        concatenate multi-scale information from the global
        and local contexts
        \item The network achieves state-of-the-art performances on
        CamVid and Cityscapes validation datasets. 
\end{itemize}

\section{Related Works}

The multi-scale problem in semantic segmentation can be described as the discrepancy in integrating the different sizes of rich spatial and channel-based semantic information of an object acquired from the global and local contexts of a network. DeepLabv3+ \cite{r9} also depends on fusing feature-based rich semantic information with the spatial information of objects in 2D scenes, just as in other encoder-decoder networks like U-Net and Segnet \cite{r17, r18}. Chen et al. \cite{r9} studied various down-sampling and up-sampling options for efficient semantic information processing and fast training and learning progress. They also experimented with the modifications of baselines, such as Xception and ResNet-101, to improve the efficiency of DeepLabv3+ \cite{r9}. The recent success of convolutional neural networks with the potential of dynamic weights and long-range dependencies \cite{r2, r4} also revealed that CNNs could be developed with additional methods, such as transform gates using the Sigmoid function \cite{r19}, as well as ViTs for multi-scale representations, which become the choice for the extra-large scale networks \cite{r20}.

Fusing the global and local semantic information in encoder-decoder architectures can become challenging due to the loss of semantic features during down-sampling and up-sampling \cite{r21, r22}. Li et al. \cite{r22} introduced global enhancement and local refinement methods that are integrated through a Context Fusion Block. Similarly, Guided Attention Inference Network (GAIN) \cite{r21} uses fully convolutional networks and CNN-based semantic segmentation architectures, such as DeepLab \cite{r9, r14}, with commonly used baseline architectures, like ResNet-101 \cite{r11, r23}.

Squeeze-and-Excitation block, SENet \cite{r24}, has also been introduced to improve the dependencies between the global and local semantic information and spatial and channel-wise features. Similarly, self-attention mechanisms are developed to acquire feature-rich information of objects at each position by aggregating features from all locations in a single sample \cite{r25}. Accordingly, Guo et al. \cite{r25} proposed external attention module using memory units by replacing the self-attention mechanisms in semantic segmentation networks. The inspirations from SENet and self-attention mechanisms also motivated SAB Net with an end-to-end semantic attention-boosting framework \cite{r26}. The applied method proposes a non-local semantic attention framework, which regularizes the difference between the non-local and local information by applying a category-wise learning weight \cite{r26}. CTNet has another alternative approach, utilizing two modules: Channel Contextual Module, for exploring the multi-scale local channel contexts, and Spatial Contextual Module, for exploring the global spatial dependency, in a tandem configuration \cite{r27}. 

In integrating the transformer architectures with self-attention mechanisms, CoTNet was developed with Contextual Transformer Block, which enables transforming and replacing each discrete convolutional operator, such as 3 × 3 convolution, with two consecutive 1 × 1 convolutions \cite{r28}. Alternatively, Ye et al. \cite{r29} introduced cross-modal self-attention that was proposed to integrate the image and language expressions as inputs as a result of the rising success of ViTs.

Attention-based feature fusion has also emerged as another solution to the challenging multi-scale problem in semantic segmentation \cite{r30}. Across feature map is used as an alternative method in dealing with small objects, which are hard to determine with their semantic features and exact location information \cite{r31}. Differently, Choe et al. \cite{r32} proposed the attention-based dropout layer, which hides the most discriminated parts from the model by utilizing a drop mask and importance map while keeping the classification accuracy.

As another alternative to attention-fusion networks, the covariance attention method was proposed by Liu et al. \cite{r33}, such that the covariance matrix maps the dependency between local and global semantic features. Similarly, the variational structured attention mechanism was proposed by Yang et al. \cite{r34} to integrate channel-based and spatial features. According to their method, the structured attention mechanism produces the tensor products of spatial attention and channel-wise attention modules \cite{r34}. The method enables the evaluation of the probabilities of latent variables that are mapped between the channel-wise and spatial attention mechanisms \cite{r34}. Hao et al. \cite{r35} proposed spatial-detail-guided context propagation by aiming to reconstruct the lost information in low-resolution global contexts using the spatial details of shallower layers in real time. Huang et al. \cite{r3} proposed CCNet as a crisscross attention network and a novel alternative to self-attention networks and attention mechanisms.

The integration of spatial information and channel-based rich semantic features is also analyzed by various novel research dealing with RGB-D networks and 3D point clouds, such as ShapeConv, S-Conv, PointMTL, and MLFNet \cite{r36, r37, r38, r39}. Respectively, this research evaluates the potential of CNN with attention-based mechanisms and explores the potential and limits of resilient baseline architectures for different tasks including 2D and 2.5D semantic segmentation.

\section{Method}

In this research, it is aimed to develop an encoder-decoder architecture with additional attention mechanisms to improve the performance of the network by regarding the multi-scale problem and fusing semantic information from different contexts. In the exploration of efficient encoder network for fast and accurate training without losing the existing progress of baseline architectures, attention-boosting gates (AbGs) and attention-boosting modules (AbMs) are designed to excite and fuse the feature-rich spatial information into the existing networks. It is found that, residual networks, pre-trained on ImageNet dataset, can be the most efficient and fastest residual networks (Fig. \ref{fig:1}) to be improved in learning most of the features in a few epochs even if it can have the capacity problems for new features. Respectively, our encoder network is advanced by introducing attention-boosting gates to the residual networks. AbGs increase the probability of exciting the equivariant feature-rich semantic information from the selected baselines with the equivalent sizes of the input and output. Thus, the feature-rich semantic information is fused with the spatial context of the encoder by AbMs, which result in a novel efficient residual network, Efficient-ResNet. To provide an efficient transition between the encoder and decoder networks of the developed architecture, dilation-based convolution layers are used.

\begin{figure*}
    \centering
    \includegraphics[width=0.95\textwidth]{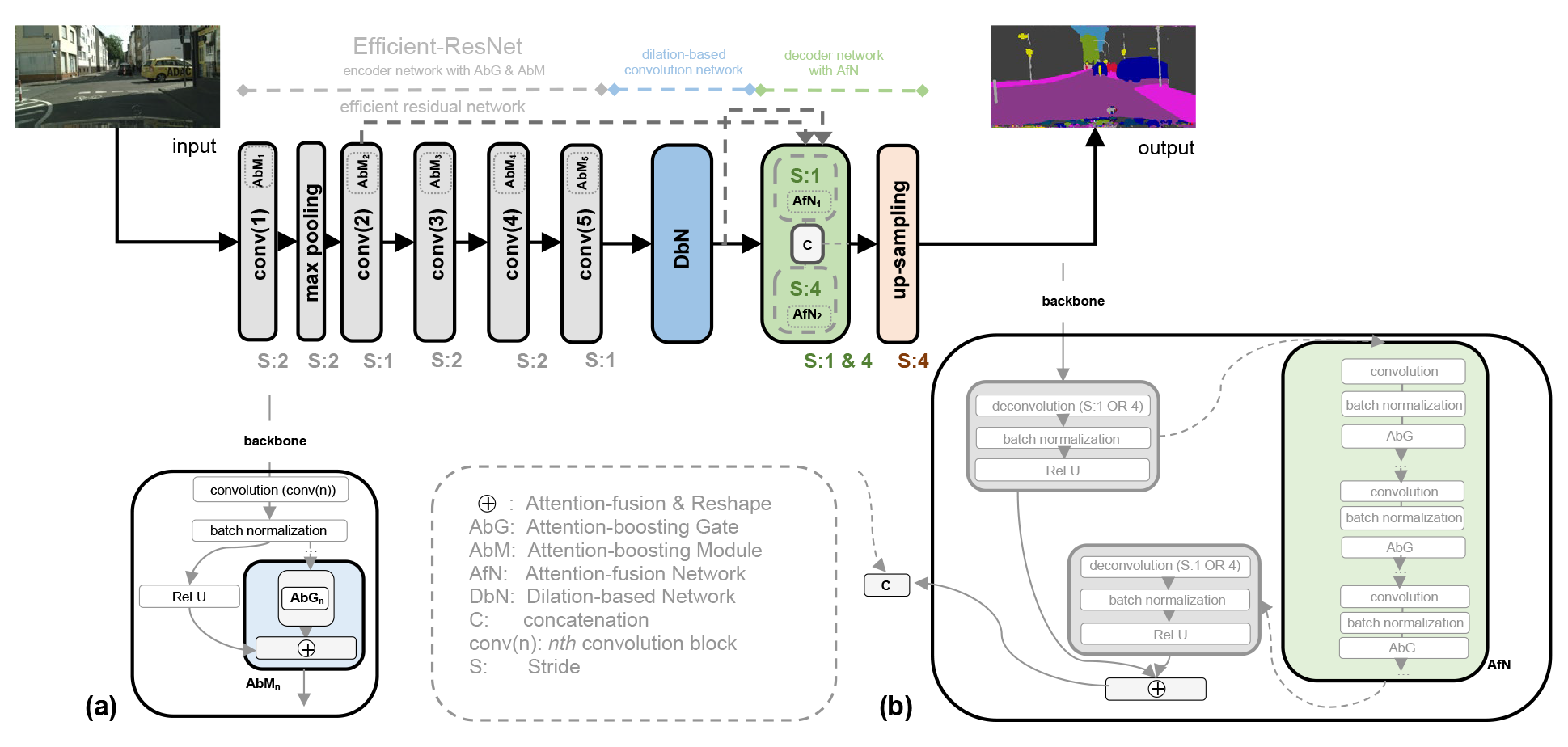}
    \caption{Schematic illustration of SERNet-Former. (a) Attention-boosting Gate (AbG) and Attention-boosting Module (AbM) are fused into the encoder part. (b) Attention-fusion Network (AfN), introduced into the decoder}
    \label{fig:2}
\end{figure*}

Prediction in semantic segmentation needs to decrease the loss during training, which is directly related to the applied loss function as well as the one-to-one conversion of the acquired data from the encoder network, transferred to the classification layer. Thus, it is also related to the efficiency of decoding the semantic information. Inspired by AbG, attention-fusion networks (AfNs) are designed to increase the efficiency in signal processing in the decoder part of our network by fusing the feature-rich spatial information from the encoder. In overcoming the capacity problems of smaller and medium residual networks, convolution layers are also added in attention-fusion networks. Respectively, skip connections are designed to have the most efficient results for fusing multi-scale feature maps in the decoder part of our network. 

Finally, the loss function and pixel classification layer of our network are set with regard to the class weights of each class in the experiment datasets and the commonly applied evaluation methods in the literature. The methods applied in improving the efficiency of our network can be briefly denoted as follows:

\begin{itemize}
    \item An efficient residual network is developed as the encoder part of our network with AbGs and AbMs
    \item Dilation-based convolution network is introduced between the encoder and decoder parts
    \item Decoder part of our network is improved by AfNs with the help of skip connections
    \item Loss function and pixel classification layer are optimized with regard to the applied evaluation methods and datasets.
\end{itemize}

\subsection{Efficient-ResNet: Encoder with attention-boosting modules}

Attention-boosting gate aims to excite channel-wise rich semantic information that may not be filtered through ReLU layers. The output of the Sigmoid function increases the possibility of producing feature-rich maps that may not be activated through residual networks. In that regard, the Sigmoid function, which is widely applied in attention networks, is selected as the operator to increase the possibility of acquiring and processing equivariant, channel-wise, and feature-based rich semantic information that could not be activated in the commonly used residual network architectures. The operation of the gate, AbG, can be iterated in Equation (\ref{eq:1}) as follows.

\begin{equation}
\label{eq:1}
    AbG_n= \sigma (i(BN(conv_n))) \times i(BN(conv_n)),
\end{equation}       

\noindent where \begin{math} i(BN(conv_n)) \end{math} denotes the output of the last convolution and the following batch normalization layers at the nth convolution block. Equation (\ref{eq:1}) defines a multiplication function, where AbG returns the feature-rich maps of the acquired channel-based semantic information. The sizes of the height and width of the output of AbG are kept equivalent to the sizes of the filtered output of the convolution layers. Accordingly, attention-boosting module, AbM, fuses the acquired and processed feature-based semantic information with the spatial context of the residual networks by element-wise addition after transforming and resizing the inputs, as illustrated in Fig. \ref{fig:2} (a). As a result of introducing the attention-boosting gates and modules into the encoder, a novel efficient residual network architecture, Efficient-ResNet, is developed (Fig. \ref{fig:2}).

AbMs are added to the baseline at the end of each \textit{n}th convolution block. It works as an attention mechanism and mathematical operator for exciting and fusing the feature-rich semantic information. It can also be introduced into different networks to acquire and process feature maps from 3D point clouds or information from RGB-D networks without re-scaling the feature-rich semantic information.

\subsection{Dilation-based separable convolution network}

Dilation-based convolution network (DbN) is applied to increase the probability of searching, recognizing, and comparing the local, channel-based, and rich semantic information between the encoder and decoder parts by decomposing the output into smaller feature maps \cite{r9} (Fig. \ref{fig:2}). Thus, the output of the encoder architecture is moved into convolutional layers, with 12, 16 and 18 dilation factors, together with the following batch normalization and ReLU layers, and they are fused together before the decoder network.

\subsection{Decoder with attention-fusion networks}

Inspired by the methods used for AbG and AbM, the efficiency of fusing global and local contexts into the decoder part is improved by attention-fusion networks. Even though the initial layers of CNNs are rich in the global context of semantic information with sharp edges and apparent shapes of objects, it is significant to transfer and reconstruct the feature-rich spatial information from the encoder during the up-sampling tasks in the decoder of the network for efficient one-to-one image processing.  

Thus, AfNs are introduced into the decoder network for transforming and fusing the semantic information with the global and local contexts from the encoder part. It is also aimed to increase the capacity of storing semantic information in the decoder regarding the limitations of smaller and simpler residual networks. Respectively, AfNs are designed and fused with additional convolution layers, as illustrated in Fig. \ref{fig:2} (b). The spatial and channel-based contextual semantic information from deconvolution layers with different strides are gathered by the concatenation of the products of AfNs via the depth concatenation layer (Fig. \ref{fig:2}). In that regard, skip connections are used to increase the efficiency of the network in acquiring the spatial information from the encoder, concatenated with the channel-based features during the up-sampling operations.

\subsection{Loss function, the classification layer and the evaluation metrics}

For calculating the performance of our semantic segmentation network, cross-entropy loss function is deployed via the pixel classification layer, as in Equation (\ref{eq:2}).

\begin{equation}
\label{eq:2}
    loss = -\sum_{x \in classes}^{C} T(x) \times {\log(Y(x))},
\end{equation}

\noindent where \textit{T} denotes the target, \textit{x} is a class in the labeled classes \textit{C} in a dataset. Thus, \textit{Y} stands for the predicted pixels. Before the experiments, the pixel classification layer is executed by calculating the class weights for each labeled class in each dataset separately. Then, cross-entropy function is deployed to use Equation \ref{eq:2} in calculating the loss between the prediction of the network and the ground truth.

\subsection{Evaluation metrics}

The exact performance in segmentation tasks is calculated by the Intersection over Union (\textit{IoU}) method deploying the \textit{Jaccard} index as in Equation (\ref{eq:3}).

\begin{equation}
\label{eq:3}
  IoU = \frac{|A \cap B|}{|A \cup B|},
\end{equation}

\noindent where A is the predicted segmentation map, and B stands for the ground truth. Thus, mean IoU (\textit{m(IoU)}) is calculated by the average \textit{IoU} over all classes \cite{r1}.

\section{Experiments and Results}
\label{others}

In this section, the selected experiment datasets and the implementation details are introduced first. The results for each open-source dataset are analyzed to be compared with other state-of-the-art models in the literature by discussing the influence of the methods applied to develop our network. Accordingly, ablation studies are conducted to analyze the contribution of each method.

\subsection{Datasets}

\paragraph{The Cambridge-driving Labelled Video Database (CamVid)} \cite{r12, r13} is one of the first scene understanding databases, and it is based on the motion-based video collections of driving scenes recorded for semantic segmentation of object classes. This database contains 701 frames with sizes of 720 by 960 pixels that were captured in five video sequences, shot via the fixed-position CCTV-style cameras mounted on a car. The densely annotated images were manually generated through 32 classes and merged into 11 classes later. The original dataset is divided into 367 training, 101 validation, and 233 test images, as most literature practiced. \phantom{\cite{r40, r41, r42, r43, r44, r45, r46, r47, r48, r49, r50, r51, r52, r53, r54, r55, r56, r57, r58, r59, r60, r61, r62}}

\begin{table*}
  \caption{Class accuracies (mIoU) of state-of-the-art methods on CamVid dataset}
  \label{Table 1}
  \centering
  \scriptsize
  \begin{tabular}{p{3.4cm} p{0.4cm} p{0.4cm} p{0.4cm} p{0.4cm} p{0.4cm} p{0.4cm} p{0.4cm} p{0.4cm} p{0.4cm} p{0.4cm} p{0.4cm} p{0.7cm}}
    \toprule 
   Method & \begin{turn}{90} Building \end{turn} & \begin{turn}{90} Tree \end{turn} & \begin{turn}{90} Sky \end{turn} & \begin{turn}{90} Car \end{turn} & \begin{turn}{90} Sign \end{turn} & \begin{turn}{90} Road \end{turn} & \begin{turn}{90} Pedestrian \end{turn} & \begin{turn}{90} Fence \end{turn} & \begin{turn}{90} Pole \end{turn} & \begin{turn}{90} Sidewalk \end{turn} & \begin{turn}{90} Bicycle \end{turn} & \begin{turn}{90} mIoU \end{turn} \\
    \midrule
    Dilation8 \cite{r41} & 82.6 &	76.2 &	89.0 & 84.0 & 46.9	& 92.2	& 56.3	& 35.8	& 23.4 & 	75.3	& 55.5	& 65.3 \\
    BiSeNet \cite{r42} & 83.0	& 75.8	& 92.0	& 83.7	& 46.5	& 94.6	& 58.8	& 53.6	& 31.9	& 81.4	& 54.0	& 68.7 \\
    VideoGCRF \cite{r43} & 86.1	& 78.3	& 91.2	& 92.2	& 63.7	& 96.4	& 67.3	& 63.0	& 34.4	& 87.8	& 66.4	& 75.2 \\
    \midrule
    Zhu et al. (SS) \cite{r44}  &  90.9 & 82.9 & 92.8 & 94.2 & 69.9 & 97.7 & 76.2 & 74.7 & 51.0 & 91.1 & 78.0 & 81.7   \\
    Zhu et al. (MS) \cite{r44} & 91.2 & 83.4 & 93.1 & 93.9 & 71.5 &97.7 & 79.2 & 76.8 & 54.7 & 91.3 & 79.7 & 82.9   \\
    \midrule
    \textbf{SERNet-Former (ours)}     & \textbf{93.0} & \textbf{88.8} & \textbf{95.1} & 91.9 & \textbf{73.9} & 97.7 & 76.4 & \textbf{83.4} & \textbf{57.3} & 90.3 & \textbf{83.1} & \textbf{84.6}    \\
    \bottomrule
  \end{tabular}
\end{table*}

\paragraph{Cityscapes} \cite{r40} is one of the most challenging datasets for the semantic segmentation of urban street scenes. It contains high-quality pixel-level annotations for 5000 images, as well as coarsely annotated 20000 images. The dataset contains diverse stereo video sequences with the sizes of 1024 by 2048 pixels, recorded during the daytime of 50 European cities visited in several months (spring, summer, and fall) with good or average weather conditions \cite{r40}. The dataset of 5000 fine annotations is divided into three sets: 2975 for training, 500 for validation, and 1525 for testing. The dataset includes semantic, instance-wise, and dense pixel annotations of 30 classes grouped into eight categories. However, most literature uses annotations with 20 classes, 19 of which are semantic labels containing objects and stuff, in addition to one additional void class for do-not-care regions.

\subsection{Implementation details and experiment results}

In the experiments on CamVid dataset, 11 classified categories are used. During the training, the images are kept to their original sizes of 720 by 960 pixels. Thus, only the single scale (SS) method is used in training and testing the dataset. The dataset split is kept similar to the commonly applied option in the literature. Respectively, the mini-batch size is set to 3, the initial learning rate is set to 0.001, and our network is trained for 80 epochs. The network is trained and tested on the arranged CamVid test dataset in MATLAB. Per-class \textit{mIoU} results with regard to the classified categories are reported in Table \ref{Table 1}. The experiment results of our network, compared with well-known state-of-the-art models, are reported in Table \ref{Table 2}. 

\begin{table}    
  \caption{Test results on CamVid dataset}
  \label{Table 2}
  \centering
  \scriptsize
  \begin{tabular}{p{2.8cm} p{0.4cm} p{0.4cm} p{2.3cm} p{0.5cm}}
    \toprule 
Method & mIoU & SS/MS & Baseline Architecture & Params (M) \\
\midrule
Dilation8 \cite{r41} & 65.3	& MS	& VGG-16	& - \\
BiSeNet \cite{r42} & 68.7	& SS	& ResNet-18	& 49.0 \\
VideoGCRF \cite{r43} & 75.2 &	SS	& ResNet-101 & - \\
CCNet \cite{r3} & 79.1	& SS	& ResNet-101	& - \\
\midrule
Zhu et al. \cite{r44}
& 81.7	& SS	& WideResNet38	& - \\
Zhu et al. \cite{r44} & 
82.9	& MS	& WideResNet38	& - \\
\midrule
RTFormer-Slim \cite{r45}& 81.4 & SS & RTFormer blocks & 4.8 \\
RTFormer-Base \cite{r45}& 82.5 & SS & RTFormer blocks & 16.8 \\
\midrule
SIW (SegFormer B5) \cite{r6, r46}& 83.7 & SS & MiT-B5 (IM-1K, MV) & 84.7 \\
\midrule
\textbf{SERNet-Former (ours)} & \textbf{84.6} & SS & \textbf{Efficient-ResNet (ours)} & 44.2 \\
    \bottomrule
  \end{tabular}
\end{table}

In overcoming the imbalanced number of instances among different categories, the weighted scores for each class in CamVid and Cityscapes datasets are analyzed first. Accordingly, the weight values are assigned in the pixel classification layer. Stochastic gradient descent with momentum (SGDM) optimizer is used as the optimization algorithm during all training sessions for both datasets. Different L2Regularization values are used throughout the training schedule to decrease losses and increase the efficiency of our network on both datasets. For both datasets, the multi-scale (MS) method is not deployed. The hardware resource of Intel® Core™ i5-6200 CPU at 2.30–2.40 GHz with 16 GB memory is used for all experiments.

Minibatch size is set to 1 during training on Cityscapes datasets to test the network for real-world scenarios with limited hardware resources. The images from the \textit{leftImg8bit} dataset are trained with instanceIDs with fine annotations, and the initial learn rate is set to 0.0005. To overcome the long period of training processes, efficient self-training methods \cite{r47}, by selecting the samples from the dataset for faster training, are deployed in the experiments. The network is initially trained through the selected 715 samples with 20 classes, including the background. Then, the developed network is further trained with 19 classes, including all samples. Thus, the network, pre-trained on CamVid dataset, is trained for 80 epochs in MATLAB for the performance evaluation. During training, the images are kept to their original sizes of 1024 by 2048 pixels. Per-class mIoU results on Cityscapes are reported in Table \ref{Table 3}. The experiment results of our network on Cityscapes datasets are reported in Table \ref{Table 4}, together with the results of well-known state-of-the-art methods.

\begin{table*}
  \caption{Per-class accuracies (mIoU) based on Cityscapes test dataset}
  \label{Table 3}
  \centering
  \tiny
\begin{tabular}{p{2cm} p{0.2cm} p{0.2cm} p{0.2cm} p{0.2cm} p{0.2cm} p{0.2cm} p{0.2cm} p{0.2cm} p{0.2cm} p{0.2cm} p{0.2cm} p{0.2cm} p{0.2cm} p{0.2cm} p{0.2cm} p{0.2cm} p{0.2cm} p{0.2cm} p{0.2cm} p{0.22cm}}
    \toprule 
Groups / Labels & \multicolumn{2}{c}{flat} & \multicolumn{3}{c}{construction}	& \multicolumn{3}{c}{object} & \multicolumn{2}{c}{nature} & sky & \multicolumn{2}{c}{person}	& \multicolumn{6}{c}{vehicle} &  \\
    \midrule
Method & \begin{turn}{90} road \end{turn} & \begin{turn}{90} sidewalk \end{turn} & \begin{turn}{90} building \end{turn} & \begin{turn}{90} wall \end{turn} & \begin{turn}{90} fence \end{turn} & \begin{turn}{90} pole \end{turn} & \begin{turn}{90} traffic light \end{turn} & \begin{turn}{90} traffic sign \end{turn} & \begin{turn}{90} vegetation \end{turn} & \begin{turn}{90} terrain \end{turn} & \begin{turn}{90} sky \end{turn} & \begin{turn}{90} person \end{turn} & \begin{turn}{90} rider \end{turn} & \begin{turn}{90} car \end{turn} & \begin{turn}{90} truck \end{turn} & \begin{turn}{90} bus \end{turn} & \begin{turn}{90} train \end{turn} & \begin{turn}{90} motorcycle \end{turn} & \begin{turn}{90} bicycle \end{turn} & \begin{turn}{90} mIoU \end{turn} \\
\midrule
Dilation10 \cite{r41} &
97.6	&79.2	&89.9	&37.3	&47.6	&53.2	&58.6	&65.2	&91.8	&69.4	&93.7	&78.9	&55.0	&93.3	&45.5	&53.4	&47.7	&52.2	&66.0	&67.1 \\
PSPNet+ \cite{r48} &
98.7	&86.9	&93.5	&58.4	&63.7	&67.7	&76.1	&80.5	&93.6	&72.2	&95.3	&86.6	&71.9	&96.2	&77.7	&91.5	&83.6	&70.8	&77.5	&81.2 \\
Gated-SCNN \cite{r49} &
98.7	&87.4	&94.2	&61.9	&64.6	&72.9	&79.6	&82.5	&94.3	&74.3	&96.2	&88.3	&74.2	&96.0	&77.2	&90.1	&87.7	&72.6	&79.4	&82.8 \\
HANet++ \cite{r50} &
98.8	&88.0	&94.2	&66.6	&64.8	&72.0	&78.2	&81.4	&94.2	&74.5	&96.1	&88.1	&75.6	&96.5	&80.3	&93.2	&86.6	&72.5	&78.7	&83.2 \\
\midrule
DeepLabv3 \cite{r14} &
98.6	&86.2	&93.5	&55.2	&63.2	&70.0	&77.1	&81.3	&93.8	&72.3	&95.9	&87.6	&73.4	&96.3	&75.1	&90.4	&85.1	&72.1	&78.3	&81.3 \\
DeepLabv3+ \cite{r9} & 98.7 & 87.0 & 93.9 & 59.5 & 63.7 & 71.4 & 78.2 & 82.2 & 94.0 & 73.0 & 95.8 & 88.0 & 73.0 & 96.4 & 78.0 & 90.9 & 83.9 & 73.8 & 78.9 & 82.1 \\
DPC \cite{r51} & 98.7 & 87.1 & 93.8 & 57.7 & 63.5 & 71.0 & 78.0 & 82.1 & 94.0 & 73.3 & 95.4 & 88.2 & 74.5 & 96.5 & 81.2 & 93.3 & 89.0 & 74.1 & 79.0 & 82.7 \\
Zhu et al. \cite{r44} & 98.8 & 87.8 & 94.2 & 64.1 & 65.0 & 72.4 & 79.0 & 82.8 & 94.2 & 74.0 & 96.1 & 88.2 & 75.4 & 96.5 & 78.8 & 94.0 & 91.6 & 73.8 & 79.0 & 83.5 \\
Panoptic DeepLab \cite{r52} & 98.8 & 88.1 & 94.5 & 68.1 & 68.1 & 74.5 & 80.5 & 83.5 & 94.2 & 74.4 & 96.1 & 89.2 & 77.1 & 96.5 & 78.9 & 91.8 & 89.1 & 76.4 & 79.3 & 84.2 \\
\midrule
Chen et al. \cite{r53} &
98.7	&87.3	&93.9	&63.8	&62.7	&70.8	&77.9	&82.2	&93.9	&72.8	&95.9	&88.2	&75.2	&96.5	&80.4	&91.6	&89.0	&73.2	&78.9	&82.8 \\
HRNet+OCR ASPP \cite{r54} & 98.8 & 88.3 & 94.3 & 66.9 & 66.7 & 73.3 & 80.2 & 83.0 & 94.2 & 74.1 & 96.0 & 88.5 & 75.8 & 96.5 & 78.5 & 91.8 & 90.1 & 73.4 & 79.3 & 83.7 \\
iFLYTEK-CV \cite{r55} & 98.8 & 88.4 & 94.4 & 68.9 & 68.9 & 73.0 & 79.7 & 83.3 & 94.3 & 74.3 & 96.0 & 88.8 & 76.3 & 96.6 & 84.0 & 94.3 & 91.7 & 74.7 & 79.3 & 84.4 \\
\midrule
SERNet-Former* (ours) & 96.8 & 76.3 & 90.0 & 57.2 & 54.6 & 52.9 & 60.5 & 66.0 & 90.9 & 64.6 & 93.9 & 79.0 & 61.6 & 93.5 & 69.7 & 85.3 & 74.7 & 59.7 & 65.6 & 73.3 \\
\textbf{SERNet-Former† (ours)} & 98.2 & 90.2 & 94.0 & 67.6 & 68.2 & 73.6 & 78.2 & 82.1 & 94.6 & 75.9 & 96.9 & 90.0 & 77.7 & 96.9 & 86.1 & 93.9 & 91.7 & 70.0 & 82.9 & 84.8 \\
    \bottomrule
    \multicolumn{21}{l}{*: ResNet-50 baseline without AbM, DbN, AfN. †: SERNet-Former with AbM, DbN, AfN} \\
  \end{tabular}
\end{table*}

\begin{table}
    \caption{Results of state-of-the-art methods on Cityscapes datasets}
    \label{Table 4} 
        \centering
        \tiny
  \begin{tabular}{p{2.7cm} p{2.4cm} p{0.4cm} p{0.4cm} p{0.4cm}}
    \toprule  
Method & Baseline architecture & \textit{mIoU}  & Params & \textit{mIoU}  \\
 & & \textit{test} dataset & (M) & \textit{val.} dataset \\
\midrule
ResNet-38 \cite{r10} &
ResNet-38 (A2, 2, val.)	&81.3	&-	&77.9 \\
PSPNet++ \cite{r48} &
ResNet-101	& 81.2 & -	&79.7 \\
PSANet+ \cite{r56} &
ResNet-101	&81.4	&-	&- \\
CCNet \cite{r3} &
ResNet-101	&81.9	&-	&80.2 \\
OCR \cite{r57} &
ResNet-101	&81.8	&10.5	&- \\
Gated-SCNN \cite{r49}&
WideResNet	&82.8	&-	&80.8 \\
HANet++ \cite{r50}&
ResNeXt-101	&83.2	&65.4	&80.3 \\
ResNeSt \cite{r58}&
ResNeSt	83.3	&70	&82.7 \\
\midrule
DeepLabv3 \cite{r14} & ResNet-101	& 81.3	& -	& 78.5 \\
DeepLabv3+ \cite{r9, r59} & Dilated-Xception-71 & 82.1 & 43.48 & 79.6 \\
Auto-DeepLab  \cite{r59} & Auto-DeepLab-L & 82.1 & 44.42 & 80.33 \\
DPC  \cite{r51} & Xception & 82.7 & 23.7 & 80.85 \\
kMaX-DeepLab \cite{r60} & ConvNeXt-L & 83.2 & 232 & 83.5 \\
Zhu et al. \cite{r44} & WideResNet38	& 83.5	& -	& 81.4 \\
Panoptic DeepLab \cite{r52} &  SWideRNet & 84.2 & 46.72 & 83.1 \\
\midrule
AdapNet++  \cite{r61} & ResNet-50 & 81.3 & 30.2 & 81.2 \\
SSMA \cite{r61} & ResNet-50 & 82.3 & 56.4 & 82.2 \\
EfficientPS \cite{r62} & EfficientNet-B5 & 84.2 & 40.89 & 82.1 \\
\midrule
SegFormer \cite{r6} & MiT-B5(IM-1K, MV) &  83.7 & 84.7 & 84.0 \\
Lawin+ \cite{r20} & Swin-L (IM-22K)	& 84.4	& 201.2	& - \\
ViT-Adapter-L \cite{r8} & ViT-Adapter-L & 85.2 & 347.9 & 85.8 \\
\midrule
HRNetV2 + OCR \cite{r54} & HRNetV2-W48 & 83.7 &  77.5 & 86.3 \\
HRNetV2+OCR+PSA \cite{r54} & HRNetV2-W48 & 84.5 & 77.5 & 86.95 \\
Tao et al.\cite{r55} & HRNet-OCR & 85.1 &	- &	- \\
HS3-Fuse \cite{r2} & HRNetV2-W18 & - &  & 81.8 \\
HS3-Fuse \cite{r2} & HRNet48-OCR-HMS & 85.7 & - & -\\
\midrule
\textbf{SERNet-Former (ours)} & \textbf{Efficient-ResNet (ours)} & 84.8 & 44.2 & \textbf{87.35}\\
    \bottomrule
  \end{tabular}
\end{table}

\subsection{Comparison with regard to state-of-the-art}

During the experiments on CamVid dataset, different baselines of DeepLabv3+, such as ResNet-18, ResNet-50, ResNet-101, Xception, and Inception-ResNet-v2, are trained in MATLAB and compared to the results of our network (Fig. \ref{fig:1}). Accordingly, ResNet-50 is found as the most efficient, due to its size and learning progress, to be modified and improved further. 

It is apparent from the results that even if MS method is not deployed, our network returns state-of-the-art results on CamVid dataset, with the help of a novel ‘efficient residual network’ and the decoder, improved with AfNs (Tables \ref{Table 1} and \ref{Table 2}). The announced accuracy for each class reveals the efficiency of our network with the applied methods when compared to the other networks, especially in recognizing the tiny objects that have less pixel area on the canvas, such as poles or bicycles, and occluding objects, such as trees and fences that can be confused with other classes, such as buildings (Table \ref{Table 1}). When compared to other methods, our network has also performed better with regard to its size and the number of parameters it uses as illustrated in Table \ref{Table 2}. The results also reveal that AbM and AfN decrease the loss in the initial training epochs successfully, increasing the actual test performance and accuracy of the segmentation network. 

\begin{figure*}
    \centering
    \includegraphics[width=1\textwidth]{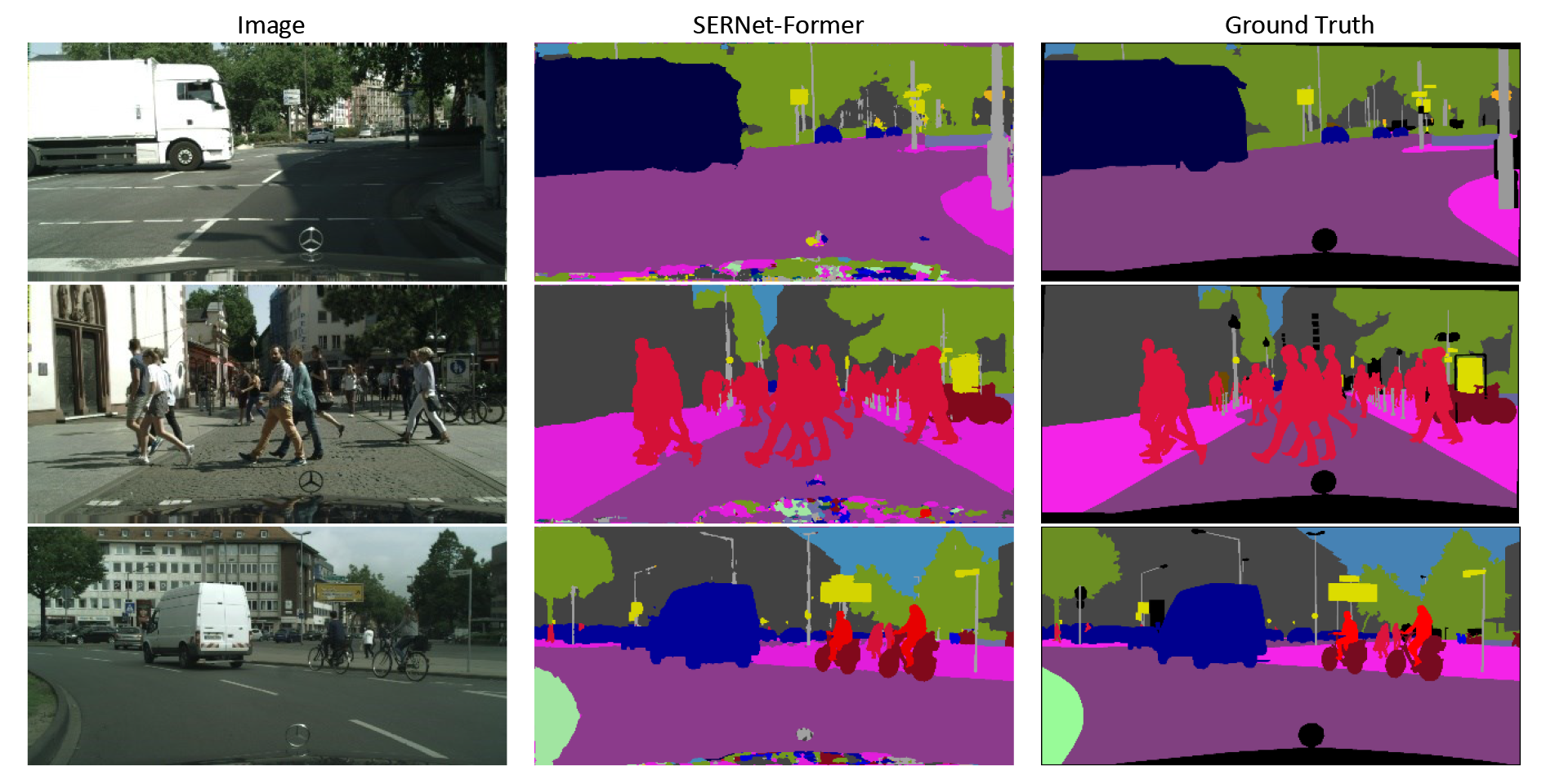}
    \caption{Examples from the test results on Cityscapes \textit{validation} dataset.}
    \label{fig:3}
\end{figure*}

Based on the results on Cityscapes test dataset, our network is successful in recognizing the objects far from the ego-vehicle (Fig. \ref{fig:3}), and especially in identifying the classes such as sidewalk, vegetation, terrain, sky, person, rider, car, truck, train, and bicycle (Table \ref{Table 3}). When considering its size and the number of parameters that are used, our network indeed performs much better than most of the state-of-the-art methods with the incomparable number of parameters, and it puts forth challenging performance results on Cityscapes validation dataset (Table \ref{Table 4}). 

On the other hand, most of the announced results of other state-of-the-art models and methods on Cityscapes datasets apply multi-scale (MS) crop sizes as well as additional, coarse datasets. In that regard, even though the results for each classified label show the considerable performance of our network, it is still hard to have a fair comparison with other state-of-the-art methods based on the results for Cityscapes datasets. Moreover, image comparisons reveal that our model can sometimes fail in predicting very sharp geometric forms that the ground truth claims even though the real-world scene also appears similar to the predicted (Fig. \ref{fig:3}), which is also left to be analyzed further.

\subsection{Ablation studies}

Ablation studies are performed on CamVid dataset during the development progress of the network. Since modules such as AbM and AfN are fused by element-wise additions, ablation studies could be done by removing the added methods and by getting the final validation of the trained network (Table \ref{Table 5}). For assessing the influence of the dilation-based network, however, a simpler network without the dilation-based convolution layers is used instead of DbN in ablation studies, and the results are reported in Table \ref{Table 5}.

AbM is designed to be introduced or removed easily to the baseline architectures. However, removing AbM from the initial convolution blocks of residual networks can change the outcomes of the network. In that regard, the efficiency of AbM can be observed best at the end of the last convolution layer of the encoder.

According to the ablation studies, the attention-boosting module (AbM5), added to the last convolution block of the encoder, augments the feature-based and channel-wise contextual semantic information of the network, and increases its efficiency by 1.656 percent (Table \ref{Table 5}). Thus, each AbM can have a share of 1.99 percent in the improvement of the network theoretically. Accordingly, DbN has a 2.79 percent share in the improved performance of the network (Table \ref{Table 5}).

\begin{table}
  \caption{Ablation studies on CamVid dataset}
  \label{Table 5}
  \centering
  \small
  \begin{tabular}{lllll}
    \toprule 
AbM5 & AfN1 & AfN2 & DbN & mIoU \\
\midrule
 & \checkmark & \checkmark & \checkmark & 81.224 (-1.656) (1.99 \%) \\
\checkmark & \checkmark & \checkmark &  & 80.562 (-2.318) (2.79 \%) \\
\checkmark &  & \checkmark & \checkmark & 78.987 (-3.893) (4.69 \%) \\ 
\checkmark & \checkmark &  & \checkmark & 75.371 (-7.509) (9 \%) \\
\midrule
\checkmark & \checkmark & \checkmark & \checkmark & 82.88 \\
    \bottomrule
  \end{tabular}
\end{table}

AfNs are designed to increase the performance of the decoder and decrease the training loss accordingly. The influences of the AfNs, which are introduced with the deconvolution layers with different strides in the decoder part of the network, are seen as extremely significant. When AfN is introduced into the following deconvolution layer with stride 1 (AfN1), it adds up to 3.893 percent to the overall test performance and improves the network with a share of 4.69 percent (Table \ref{Table 5}). When AfN is fused into the decoder with the following deconvolution layer with stride 4 (AfN2), it improves the network’s performance by more than 7.5 percent. Based on the earlier test results (Table \ref{Table 5}), AfN2 contributes to the improvement of the network by 9 percent as it also processes the spatial information with different sizes and contexts fused into the network. 

\section{Conclusion}

Regarding the multi-scale problem of fusing semantic information from different contexts with different sizes, a novel and efficient residual network is developed in an encoder-decoder architecture. Attention-boosting gates and attention-fusion networks, in which the Sigmoid function is used to increase the possibility of activating the channel-based feature maps, improve the efficiency of the network with the rising performance by fusing semantic information from the local and global contexts.

Efficient-ResNet is still open to be improved and tested on the challenging datasets for classification tasks as the encoder of SERNet-Former. The decoder part of our network can also still be modified with AfNs. Additional methods that are not applied during experiments, such as multi-scale (MS) crop sizes of images as well as additional coarse datasets that most literature applies, can also improve the results of our network. Moreover, SERNet-Former is developed as having potential for different tasks using feature maps from RGB-D networks and 3D point clouds, and it can be tested for real-time segmentation tasks and real-world applications with limited hardware resources, which also remain as future work. Respectively, we hope that our methods in developing Efficient-ResNet and SERNet-Former inspire many researchers to develop novel and efficient state-of-the-art deep learning architectures by enlarging the encoder and decoder parts of larger residual networks and CNN with additional methods.

{
    \small
    \bibliographystyle{unsrt}
    \bibliography{bib}
}


\end{document}